\documentclass[sigconf]{acmart}

\usepackage{graphicx}
\usepackage{latexsym}
\usepackage[disable]{todonotes}
\usepackage{dblfloatfix}
\usepackage{algpseudocode}
\usepackage{algorithm}
\usepackage{multirow}
\usepackage{subcaption}
\usepackage[inline]{enumitem}[(1)]
\algnewcommand\algorithmicforeach{\textbf{for each}}
\algdef{S}[FOR]{ForEach}[1]{\algorithmicforeach\ #1\ \algorithmicdo}
\algdef{SE}[DOWHILE]{Do}{doWhile}{\algorithmicdo}[1]{\algorithmicwhile\ #1}

\setcopyright{none}
\renewcommand\footnotetextcopyrightpermission[1]{} 

\begin{document}

\pagestyle{empty}
\title{A data-science pipeline to enable the Interpretability of Many-Objective Feature Selection}

\author{Uchechukwu F. Njoku}
  \affiliation{
  \institution{Universitat Polit{\`e}cnica de Catalunya}
  \city{Barcelona}
  \country{Spain}}
  \affiliation{
  \institution{Universit{\'e} Libre de Bruxelles}
  \city{Brussels}
  \country{Belgium}}
\email{uchechukwu.fortune.njoku@upc.edu}

\author{Alberto Abelló}
\affiliation{
  \institution{Universitat Polit{\`e}cnica de Catalunya}
  \city{Barcelona}
  \country{Spain}}
\email{alberto.abello@upc.edu}

\author{Besim Bilalli}
\affiliation{
  \institution{Universitat Polit{\`e}cnica de Catalunya}
  \city{Barcelona}
  \country{Spain}}
\email{besim.bilalli@upc.edu}

\author{Gianluca Bontempi}
\affiliation{
  \institution{Universit{\'e} Libre de Bruxelles}
  \city{Brussels}
  \country{Belgium}}
\email{gianluca.bontempi@ulb.be}

\begin{abstract}
Many-Objective Feature Selection (MOFS) approaches use four or more objectives to determine the relevance of a subset of features in a supervised learning task. As a consequence, MOFS typically returns a large set of non-dominated solutions, which have to be assessed by the data scientist in order to proceed with the final choice. 
Given the multi-variate nature of the assessment, which may include 
criteria (e.g. fairness) not related to predictive accuracy,  this step is often not straightforward and suffers from the lack of existing tools.
%
%
For instance, it is common to make use of a tabular presentation of the solutions, which provide little information about the trade-offs and the  relations between criteria over the set of solutions. 

This paper proposes an original methodology to support data scientists in the interpretation and comparison of the MOFS outcome by combining post-processing and visualisation of the set of solutions. The methodology supports the data scientist in the selection of an optimal feature subset by providing her with high-level information at three different levels: objectives, solutions, and individual features. 

The methodology is experimentally assessed on two feature selection tasks adopting a GA-based MOFS with six objectives (number of selected features, balanced accuracy, F1-Score, variance inflation factor, statistical parity, and equalised odds). The results show the added value of the methodology in the selection of the final subset of features. 

\end{abstract}


\maketitle
\section{Introduction} \label{intro}
The increasing availability of sensors and data sources in the \textit{big data} era confronts data scientists with datasets of huge size and dimensionality. 
Effective Feature Selection (FS) is then more and more important
in order to extract \textit{valuable information} by identifying relevant features and discarding irrelevant or redundant ones.

Given a supervised task with $m$ input features, the number of possible subsets is exponential (exactly $2^m-2$, excluding the empty and full set). FS denotes the techniques to find one or more subsets of features (in the following, also called solutions) containing the most relevant ones. Most definitions of relevance of features rely on accuracy or information aspects~\cite{al2020approaches}. However, we should consider relevance in a broader sense, as numerous other objectives could be considered: subset size, predictive accuracy, F1-score,  Area Under the ROC Curve (AUC), hamming loss, clustering quality, feature reconstruction error \cite{al2020approaches}, and fairness measures \cite{salazar2021automated}. 

The set of objectives is highly dependent on the task and the goal of the data analysis. For example, if we aim at predicting hospital readmission of patients with Diabetes \cite{diabetes}, the objectives could include  generalisation-related measures (e.g., size of the subset, predictive performance, redundancy) as well as fairness criteria. The inclusion of  fairness objectives is  more and more required in the application of artificial intelligence (AI) to domains involving confidential information and having a potential impact on citizens' rights and welfare~\cite{le2022survey} (e.g., healthcare, law, education).

Moving from a single to several objectives requires the adoption of an FS strategy able to account for the trade-offs among the considered objectives. In particular, given the impossibility of selecting an optimal solution, many-objective FS methods return a number of non-dominated solutions, among which the data scientist is expected to select the most convenient one. In the case of two objectives, the set of solutions can be illustrated with 2D line plots~\cite{hancer2018pareto}, scatter plots \cite{zhang2019cost}, bar charts \cite{grzyb2021application}, or simply tables \cite{liu2021interactive}. With three objectives, it becomes more complex to present the solutions on a single graph, requiring 3-D scatter plots \cite{xue2021multi}, multiple charts \cite{barbiero2019novel} or tables \cite{zhou2021evolutionary}. When we move to four or more objectives, 
it is even harder for the data analyst to extract some useful insight (e.g., dominance relationships between solution \cite{shavazipour2021visualizations}) from graphical representations
(e.g., the parallel coordinate plot)  which are typically over-cluttered. 

The few current works on MOFS use a set of tables to present the various aspects of the set of solutions \cite{zhang2020many} or adopt a univariate score, based on classifier accuracy \cite{jha2021incorporation} or specific measures \cite{zhang2020many,rodrigues2020multi}. As a consequence, current MOFS solutions are either based on a large number of descriptive tables or rely 
on a single measure hiding the multi-variate nature of the objective space.  Thus, after the massive computational effort
spent in generating a number of interesting solutions, the final selection does not take advantage of  any interpretable and user-friendly exploration mechanism~\cite{zacharias2022designing}.
 \begin{figure*}[ht]
\centerline{\includegraphics[height=1.5in]{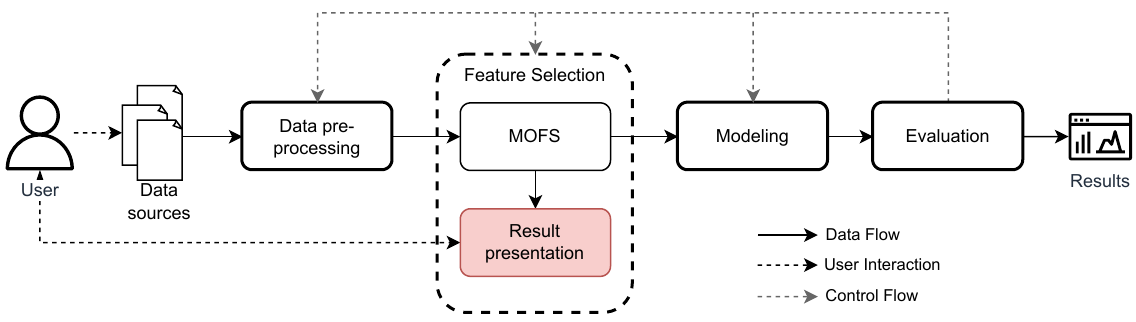}}
\caption{An end-to-end data science process. } \label{framework}
\end{figure*}

In this work, we propose a novel methodology to enable an effective  exploration of the MOFS set of solutions by providing the data scientist 
with useful visual representations and post-processing of those solutions. In particular, we extend the classical data science pipeline with a MOFS interpretation module (Figure~\ref{framework}), including the visualisation and the presentation of the set of solutions
from three different perspectives: objectives, subsets of features (a.k.a., solutions) and individual ones.  We followed a \emph{concept analysis} research method to systematically identify the different functionalities of the module that are detailed in Section~\ref{sec:methodology}.

To illustrate the added value of the approach, we consider two classification tasks with six objectives (i.e., the subset size, balanced accuracy, F1-score, Variance Inflation Factor – V.I.F, statistical parity, and equalised odds). We use a Genetic Algorithm (GA), namely NSGA-III \cite{deb2013evolutionary}, as a search strategy for generating the set of solutions with Naive Bayes (NB) and Logistic Regression (LR) as classifiers.\footnote{We applied our methodology to eight more datasets with four objectives in MOFS. For lack of space, we do not discuss the results in this paper. However, they are available on Github \url{https://github.com/F-U-Njoku/many-objective-fs-nsgaiii}.}

In summary, the contributions of this work are:
\begin{enumerate}[label=(\alph*)]
    \item We propose a methodology to visualise, interpret, assess and compare the set of MOFS solutions from three different perspectives:  objectives, solutions, and individual features.
    \item We show the feasibility and added value  of our methodology through an experimental assessment considering two common fairness benchmark datasets.
    \begin{itemize}
        \item Also, we evaluate the robustness of the outcomes of our methodology according to the dataset characteristics as well as the weights of the objectives. 
    \end{itemize}
\end{enumerate}

\section{Background} \label{background}
 We present here some useful background notions about FS approaches, many-objective optimisation and the related search strategies.

\subsection{Single-objective FS strategies}
Single-objective FS methods are categorised into \textit{filter}, \textit{wrapper}, \textit{embedded}, and \textit{hybrid} \cite{jovic2015review} according to the role of the learning algorithm.
\textit{Filter} methods do not rely on any learning algorithm and the relevance of a subset of features is measured by a model agnostic measure of its association (e.g., mutual information) with the target~\cite{njoku2022impact}. \textit{Wrapper} methods rely on the optimisation of a learner-dependent accuracy cost function over the space of solutions. Their solution is model-specific and often better than filter methods for the learning algorithm used \cite{njoku2023wrapper}. \textit{Embedded} methods are learning algorithms that do FS intrinsically; this includes tree-based algorithms. Like wrapper methods, the solution is not model agnostic but obviously tailored to the used learning algorithm. Lastly, \textit{hybrid} methods combine two or more of the aforementioned methods in order to gain the advantages they each bring.

\subsection{Multi-objective Optimisation Approaches}
Finding optimal solutions by simultaneously considering two or more, often conflicting objectives can be done by either the \textit{Scalarisation} (preference-based) or \textit{Pareto} approach \cite{deb2014multi}.

The \textit{scalarisation} approach simply merges the considered objectives into a  single one, while the \textit{Pareto} approach dynamically finds near Pareto optimal solutions.\footnote{Pareto optimal solutions are solutions for which no other solution exists that improves one of the objectives without deteriorating some of the other objectives.} The \textit{Pareto} approach finds multiple solutions with different trade-offs of the considered objectives; one of these solutions is then chosen as final. This approach typically works with heuristics such as GAs. 



\subsubsection{NSGA-III}  \label{nsgaii}
\begin{figure}[ht]
\centerline{\includegraphics[height=2.5in]{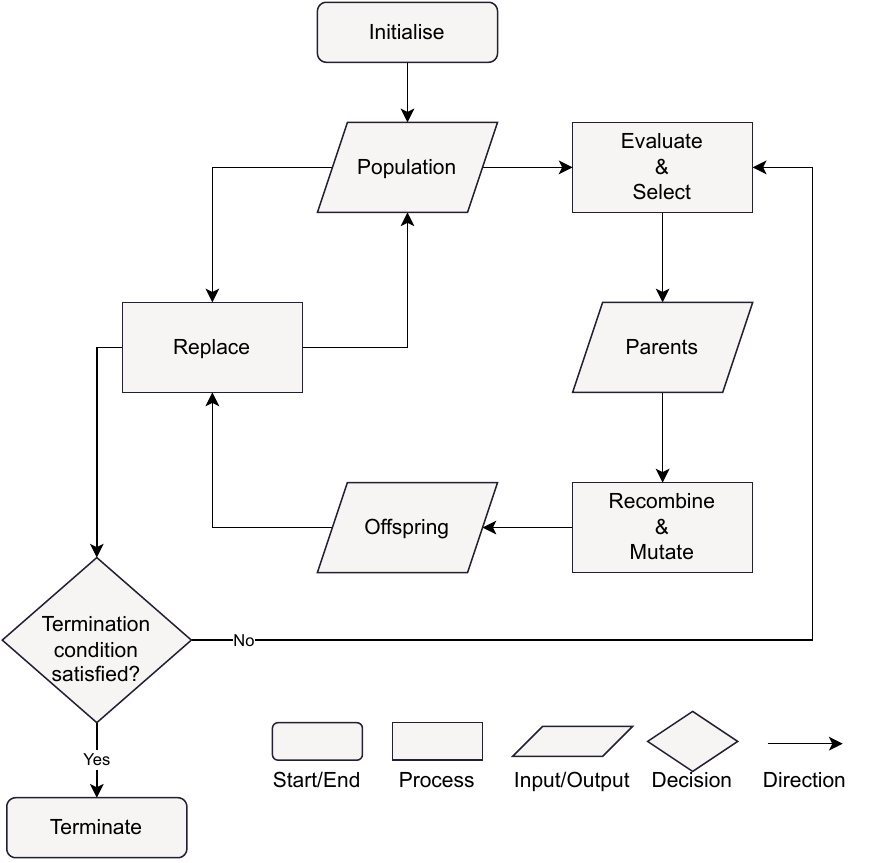}}
\caption{A flowchart of the NSGA-III process. } \label{nsgachart}
\end{figure}

NSGA-III is a pareto and reference-based elitist GA for many-objective problems \cite{deb2013evolutionary}. It uses well-spread out reference points to maintain population diversity and is elitist because of being designed to preserve the set of best individuals at each iteration. The NSGA-III algorithm begins by initialising a set of candidate solutions (i.e., population) and continues to optimise towards better solutions until a set termination criterion is satisfied as shown in Figure~\ref{nsgachart}. We use it for our experiments.

\begin{figure*}[ht]
\centerline{\includegraphics[width=1\textwidth]{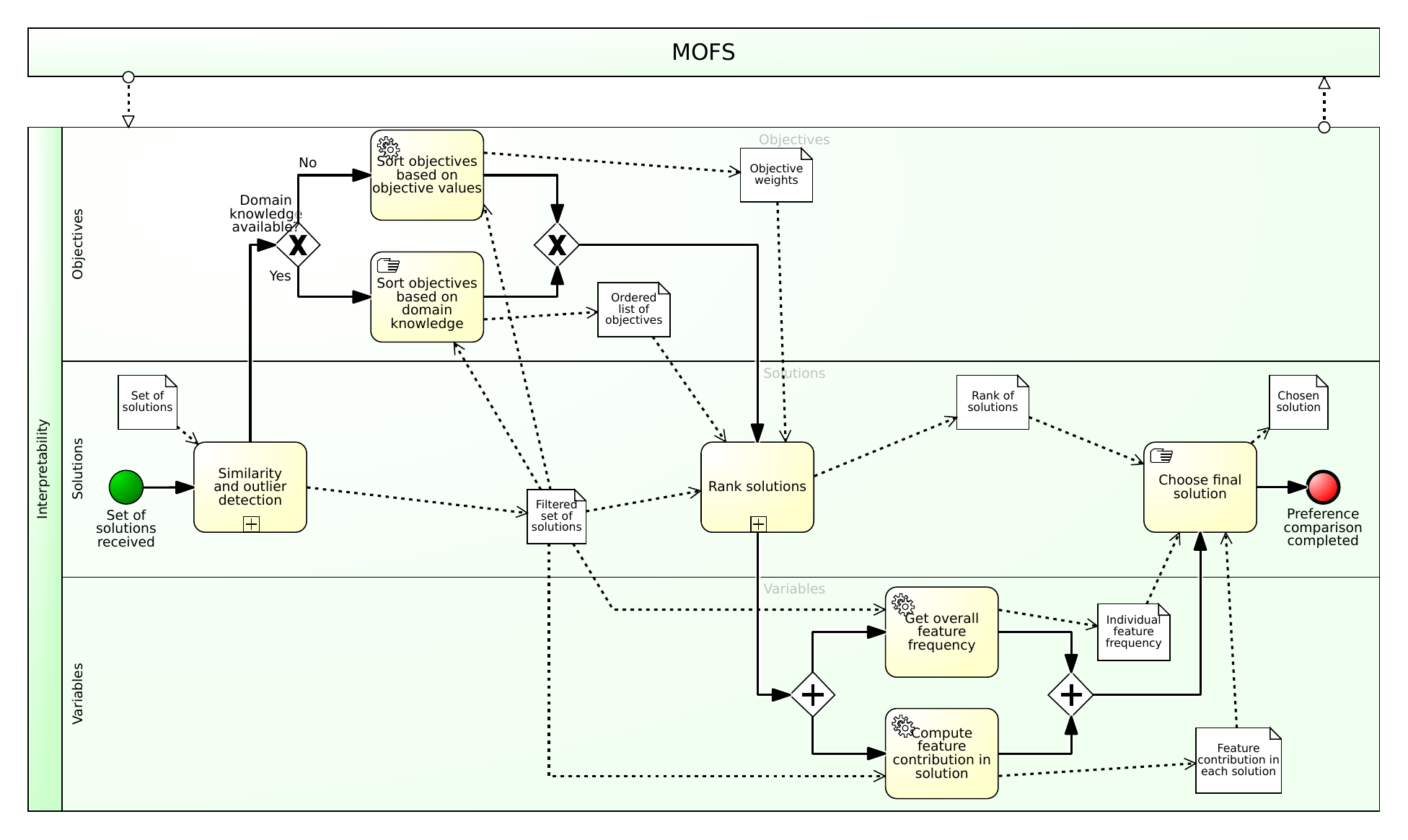}}
\caption{A methodology for the interpretation of MOFS results.} \label{method_fin}
\end{figure*}

\section{Methodology}\label{sec:methodology}
FS sits at the heart of the data science process as shown in Figure~\ref{framework}. The MOFS method produces a set of results that needs to be presented in an interpretable way to facilitate alternative comparison so that the decision maker can choose a final solution to use for further analysis. Similar to other many-objective problems, the goal is to provide the decision maker with an overall view of the results, along with details for evaluating alternatives and comparing preferences \cite{korhonen2008visualization}. We therefore propose a methodology, depicted in Figure~\ref{method_fin}, that enables the interpretability of MOFS results.

For an overall view of the results, we consider three viewpoints in our methodology: \textit{objectives}, \textit{solutions}, and \textit{features} (a.k.a., \textit{variables}). Each step of our methodology corresponds to one of these viewpoints as shown by the lanes in Figure~\ref{method_fin}, and in the following, we briefly present each step, as well as potential techniques to implement it (the final choice being case dependent).
\begin{enumerate}[noitemsep]
    \item \textbf{Detect similar/outlier solutions. }Given a set of solutions, some of them might be similar with respect to the objectives. Oppositely, there could also be some solutions substantially different from the rest. The first step is to detect these highly similar or different solutions, which is not a trivial task and could be achieved through various approaches. One way to do this is to cluster the solutions. Another way is via  pairwise visualisations. Moreover, we can also achieve it through multidimensional scaling (MDS) \cite{saeed2018survey} methods such as Principal Component Analysis (PCA) and other similarity measures such as Jaccard index \cite{verma2020comparative}. The fitting approach depends on the characteristics of solutions and the know-how of the decision maker. Upon identification of these similar/different solutions, the user can either reduce the number of candidates by discarding some or get them simply labelled and proceed.
    \item \textbf{Sort objectives.} The next step is to order the objectives. Often, the objectives considered do not have equal importance and this should be taken into account in understanding the solutions and choosing a final one. Although the decision maker might not be able to assign specific weights to the objectives, he/she can order them by their relevance; which requires domain knowledge. When such domain knowledge is not available, we can use various statistical and numerical methods to sort and generate weights for the objectives (e.g., the variance of the values in the solutions for each objective or the entropy weight \cite{oluah2020selection}). This order or weights of the objectives is used in the ranking of the solutions as well as in determining the pairs of objectives to consider during the pairwise comparison. The outcome of this task is either a sorted list or some weights of the objectives.
    \item \textbf{Rank solutions. }With the similarity of solutions handled and the objectives ordered, we can proceed to rank the solutions. We can achieve this by using Multiple-Criteria Decision-Making (MCDM) methods \cite{mardani2015multiple, alvarez2021multiple} such as the Technique for Order of Preference by Similarity to Ideal Solution (TOPSIS) \cite{oluah2020selection}. Alternatively, we could use a graph theory and matrix approach \cite{Rao2007}, too. Solutions can also be first ranked by pairs of objectives followed by summarising these pairwise ranks. As with detecting similar solutions, the fitting approach for ranking depends on the characteristics of solutions available and the know-how of the decision maker. The outcome of this task is a rank of solutions. 
    \item \textbf{Get feature frequency. }So far, we have considered the objectives and solutions viewpoints. This step corresponds to the more specific variable viewpoint and focuses on the individual features and their relevance according to the set of solutions. Particularly, the relevance of individual features is summarised by their appearance in the set of solutions. The more a feature appears in the set of solutions, the more relevant it is. The outcome of this task is the frequency of individual features in the set of solutions.
    \item \textbf{Compute feature contribution. }  Also under the variable viewpoint, this task focuses on individual features and their contribution to the task at hand (e.g., classification). Considering the classification task, each feature discriminates the target classes differently, which means they contribute differently. Understanding how the selected features discriminate the target classes (in case of dealing with a classification problem) is useful in making the choice of the final solution. The goal at this level is to have features that contribute to predicting all the classes in the target. To measure the contribution of the features, we can use information theory methods such as Gini and information gain or explainability measures such as SHAP \cite{marcilio2020explanations}, which shows how each feature discriminates the classes.
    \item \textbf{Choose final solution. }With the outcomes of the previous steps, the decision maker is equipped with sufficient insight about the objectives, solutions, and features, to choose one solution for further analysis. If needed, the MOFS process can be repeated with different settings to fine tune the final set of solutions. 
\end{enumerate}
\section{Experiments} \label{experiments}
To assess the efficacy of our proposed methodology, we apply it to the set of solutions of a hybrid MOFS method on two commonly used fairness datasets. The MOFS method uses filter methods to measure subset size and VIF objectives, and wrapper methods to measure balanced accuracy, F1-score, statistical parity, and equalised odds objectives. It also uses NSGA-III which is a pareto-based optimisation strategy and a population-based search method. In this section, we present the MOFS method and datasets used in this work as well as the execution details for reproducibility.
\subsection{MOFS implementation}
We describe the starting point (i.e., initialisation of the first population), search strategy, feature subset evaluation (i.e., the objectives), and the termination criterion for the MOFS used in this work.
\subsubsection{Starting Point}  \label{startpt}
We begin with a population of one-sized candidate solutions, where each feature is selected at least once. The population size $p$ is set greater than the overall number of features $m$. So, $p = m+1$ if $m$ is odd. Otherwise, $p = m+2$.

\subsubsection{Search Strategy and Subsets Evaluation}
We use NSGA-III with several parameters as the search strategy for MOFS. In particular, we set the mutation probability to $1/m$ and crossover probability to $1$.

Knowing that one measure of relevance for a feature subset is insufficient, we define subset evaluation by up to six objectives:
\begin{enumerate}[noitemsep]
	\item The \textbf{subset size} is a fundamental objective of FS and the goal is to minimise it ($\downarrow$).
	\item \textbf{Balanced accuracy} measures accurate predictions. It is a commonly used predictive performance metric in classification problems and we aim to maximise it ($\uparrow$).
	\item \textbf{F1-Score}, which is the harmonic mean of precision and recall \cite{hicks2022evaluation}, is another predictive performance measure which we maximise ($\uparrow$).
	\item \textbf{Variance Inflation Factor (VIF)} measures multicollinearity between features \cite{cheng2022variable}. A high VIF implies redundancy, which we minimise ($\downarrow$).
     \item \textbf{Statistical parity} checks if classifier predictions are void of group membership sentiments \cite{le2022survey}, and we aim to maximise it ($\uparrow$).
     \item \textbf{Equalised odds} measures if the classifier performs equally well across sensitive groups \cite{le2022survey}, and we aim to maximise it ($\uparrow$).
\end{enumerate} 
The goal therefore, is to find subsets of features with the least possible number of relevant ones that are minimally correlated and yield fair and satisfactory predictions. We used NB and LR classifiers to measure the accuracy, F1-score, statistical parity, and equalised odds. 
\subsubsection{Termination Condition}
In this work, we set the termination criterion to a given maximum number of evaluations performed, which is set up to $2p^2$. When this condition is satisfied, the non-dominated \cite{deb2013evolutionary} solutions in the current population are returned. 
\subsection{Datasets}
We used two fairness benchmark datasets in our experiments, whose details are shown in Table~\ref{datatab}. 
In the Diabetes\footnote{https://doi.org/10.24432/C5230J} dataset, the task is to predict hospital readmission of patients with Diabetes, while in the German credit\footnote{https://doi.org/10.24432/C5QG88} dataset, we try to predict the credit risk of bank account holders.
\begin{table}[ht]
\begin{center}
{\caption{Dataset properties.}\label{datatab}}
\resizebox{\columnwidth}{!}{
\begin{tabular}{crrrrrr}
\toprule
\textbf{Name} & \textbf{\#Features}& \textbf{\#Instances} & \textbf{\#Classes} & \textbf{Sensitive} & \textbf{Pop. Size} & \textbf{Max. Evals.}\\
\midrule
Diabetes      &    $57$     & $101,766$      &   $2$    &  Gender        &  $58$  &  $6728$  \\
German credit     & $21$         & $1000$       & $2$     & Sex        & $22$  & $968$ \\
\bottomrule
\end{tabular}}
\end{center}
\end{table}
\subsection{Execution}
We implemented the experiments using Python~3 and particularly, the \textit{jMetalPy}\footnote{https://github.com/jMetal/jMetalPy} library implementation of NSGA-III \cite{benitez2019jmetalpy}. The source code for our experiments is available on Github.\footnote{https://github.com/F-U-Njoku/many-objective-fs-nsgaiii} We executed the experiments on one node in the Dragon2 cluster; one of the 6 clusters provided by \textit{Consortium des Équipements de Calcul Intensif}\footnote{https://www.ceci-hpc.be}- a consortium of high-performance computing centres in five Belgian universities. The node has two Intel Skylake 16-cores Xeon 6142 processors at 2.6 GHz.
\subsection{\textbf{Methodology Implementation}} 
We present below the details to instantiate our proposed methodology.
\begin{enumerate}[noitemsep]
\item \textbf{Detect similar/outlier solutions.} The solutions for each dataset are clustered using K-Means with $3$ clusters (chosen based on the elbow criterion method). We then project the solutions to 2-D using principal component analysis and plot them on a scatter plot showing the clusters of similar solutions as well as outliers.

\item \textbf{Sort objectives.} To sort the objectives, we use their values in the solutions to compute the weight for each objective as $r/\sigma$ based on the range $r$ (i.e., $maximumValue - minimumValue$) and standard deviation $\sigma$. Higher weights mean more variance in the objective values which differentiates the solutions better. Therefore, the weights reflects the importance of each objective compared to the others.

\item \textbf{Rank solutions.} We use TOPSIS MCDM method \cite{oluah2020selection}. TOPSIS method requires the weight of the objectives and hence, we use the weights obtained earlier. With TOPSIS, the best solution should have the shortest euclidean distance from the positive ideal solution ($S_i^+$) and the farthest from the negative ideal solution ($S_i^-$).\footnote{For each objective, the positive and negative ideal solutions are the best and worst values attained in the solutions respectively.} Beginning with a normalised matrix of the solutions $\bar{X}$, values in each column -representing the objectives- are multiplied by the corresponding objective weight, ($\bar{X_j} * w_j$). Next, for each solution, we compute its euclidean distance from the positive and negative ideal values for each objective using the weighted normalised matrix. The final performance score is derived by Equation \ref{topsis}.
\begin{equation}\label{topsis}
    PS = \frac{S_i^-}{S_i^+ + S_i^-}.
\end{equation}
\item \textbf{Feature frequency.} The frequency of the features is computed by counting the occurrence of each feature in the set of solutions. 
\item \textbf{Feature contribution.} We use SHAP to measure the global contribution of each selected feature. 
\item \textbf{Final solution. }Based on the ranks, feature frequency and contribution, we discuss a possible final solution for each dataset.
\end{enumerate}
\section{Results and Discussion} \label{results}
Through our methodology, we are able to identify the most impactful objectives, solutions, and features from both datasets; something not possible through mere tabular presentation or even parallel coordinate plots.
Indeed, for each of both datasets used in our experiments, we scrutinise and interpret its set of solutions which we present below. 

\subsection{Diabetes}
For this dataset, we try to predict hospital readmission of patients with Diabetes using the NB classifier, and the sensitive feature on which we strive for fairness is \textit{Gender}.  The MOFS produced $52$ non-dominated solutions, and the ranges of values for the objectives are as follows: subset size $[7, 19]$, balanced accuracy $[0.5049, 0.5422]$, F1-score $[0.0456, 0.1790]$, VIF $[0.0093, \infty]$, statistical parity $[0.8432, 0.9246]$, and equalised odds $[0.7130, 0.8403]$. For VIF, we replace $\infty$ with $10$ which is about ten times larger than the preceding value. 
 
 \begin{figure}[ht]
	\centering
		\centering
		\includegraphics[height=1.5in]{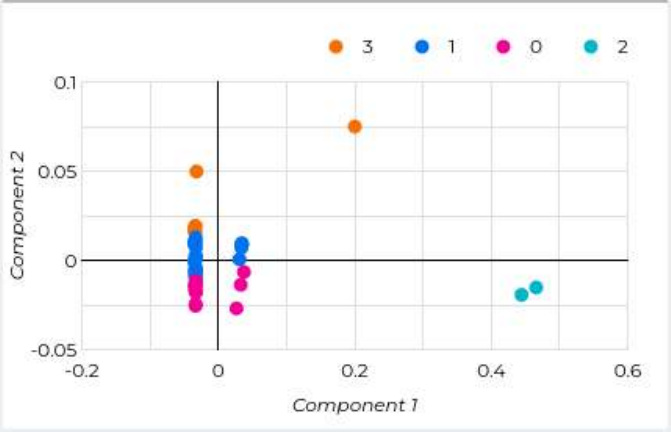}
		\caption{PCA with 2 components for Diabetes data.}
		\label{diab_cluster}
\end{figure}
 
Firstly, we look for similar and outlier solutions using K-Means clustering with $3$ clusters. We project the solutions into 2-D using PCA to visualise the clusters as shown in Figure~\ref{diab_cluster}.
We observe that cluster~$2$ corresponds to the solutions with an $\infty$ VIF while the remaining are split between clusters~$0$, ~$1$ and~$3$. Proceeding with all solutions, we label each one with a cluster ID so we can examine the correlation between how solutions are clustered and ranked. However, based on the clusters of solutions, the decision maker could discard solutions that are highly similar.

\begin{table}[ht]
\caption{Weight and rank of objectives for Diabetes dataset.}
\label{tab:diabetes_obj_rank}
\begin{tabular}{lrr}
\toprule
\textbf{Objective} & \multicolumn{1}{l}{\textbf{Weight}} & \multicolumn{1}{l}{\textbf{Rank}} \\
\midrule
$\downarrow$Subset size        & 0.1973 & 1 \\
$\downarrow$VIF                & 0.1904 & 2 \\
$\uparrow$F1 Score           & 0.165  & 3 \\
$\uparrow$Balanced accuracy  & 0.1588 & 4 \\
$\uparrow$Statistical parity & 0.1517 & 5 \\
$\uparrow$Equalised odds     & 0.1367 & 6 \\
\bottomrule
\end{tabular}
\end{table}

The variance of values for each objective is different and this should reflect in the final decision making. Using the method described in Section~\ref{experiments}, we compute the weight and rank of each objective. As shown in Table~\ref{tab:diabetes_obj_rank}, the subset size takes the highest weight and rank while the equalised odds comes last. This is a reflection of the variance in the values of the objectives in the solutions.

\begin{table}[ht]
\caption{Comparison of TOPSIS top solution to baseline for Diabetes dataset.}
\label{tab:topsis_baseline}
\begin{tabular}{lrr}
\toprule
\textbf{Objective} & \multicolumn{1}{l}{\textbf{TOPSIS - 1}} & \multicolumn{1}{l}{\textbf{Baseline}} \\
\midrule
$\downarrow$Subset size        & \textbf{10}     & 56                      \\
$\uparrow$Balanced accuracy  & 0.5267 & \textbf{0.5538}                  \\
$\uparrow$F1 Score           & 0.1399 & \textbf{0.2070}                   \\
$\downarrow$VIF              & \textbf{0.0206}   & $\infty$                  \\
$\uparrow$Statistical parity & 0.8498 & \textbf{0.8715}                  \\
$\uparrow$Equalised odds     & 0.7775   & \textbf{0.8204}                  \\
\bottomrule
\end{tabular}
\end{table}

Using the weights obtained for each objective, we rank the solutions using the TOPSIS method. The solution with $10$ features emerged as the highest ranked one with the respective objective values shown in Table~\ref{tab:topsis_baseline}. When aiming for a fair model, a simplistic approach would to discard the sensitive attribute (i.e, gender) totally from the analysis. Comparing the best result from TOPSIS rank with this simplistic approach as baseline shows that including fairness  metrics (i.e., statistical parity and equalised odds) for feature selection leads to a subset of features with comparable scores for those metrics. A number of other solutions achieved better performance than the baseline. For example, the solution ranked sixth with $11$ features, achieved statistical parity and equalised odds of $0.8892$ and $0.8249$ respectively. 

Finally, focusing on features selected at least once, we evaluate their individual relevance based on their mean SHAP value as well as the number of times they were selected. According to the SHAP values, features \textit{had\_emergency} and \textit{had\_inpatient\_days}  which indicates if a patient was admitted in the previous year are the most relevant. By frequency, the \textit{primary\_diagnosis} and \textit{had\_inpatient\_days} features were selected $51$ and $45$ times respectively; out of the $52$ solutions. All three features are present on the TOPSIS top ranked solution. Furthermore, our results show that for this dataset, there is no association between the solutions clusters and ranks.

\begin{figure}[ht]
	\centering
	\includegraphics[height=1.5in]{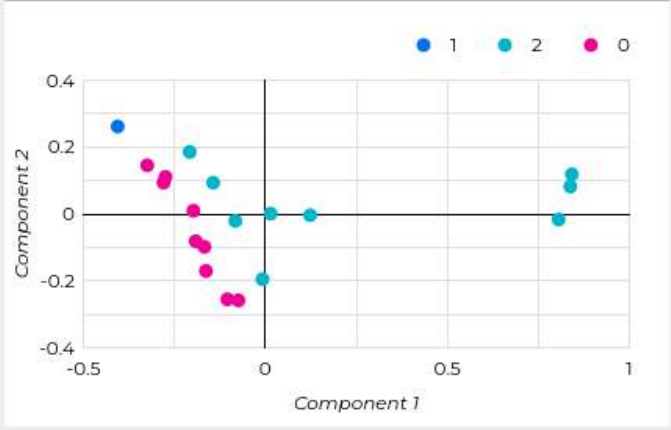}
    \caption{PCA with 2 components for German credit data.}
	\vspace{-2mm}
	\label{fig:german_pca}
\end{figure}

\subsection{German credit} 
The task for this dataset is to predict the credit risk of bank account holders using a LR classifier, having \textit{famges (Sex)} as a sensitive feature. MOFS produced $19$ non-dominated solutions where the range of values for each objective is: subset size $[1, 10]$, balanced accuracy $[0.5000, 0.6607]$, F1-score $[0.7776, 0.8305]$, VIF $[0, 16.668]$, statistical parity $[0.7488, 1]$, and equalised odds $[0.2122, 0.8000]$. To begin interpreting the results, we group them with K-Means into $3$ clusters. In Figure~\ref{fig:german_pca}, we observe quite a spread of the solutions which are mostly split between cluster $0$ and $2$, and only one solution is found into cluster $1$ (this is the solution with a VIF of $16.67$, much higher than the others). Also, solutions in the same cluster are closely ranked with all $9$ solutions in cluster $2$ as the top 9 ranked solutions. The only solution in cluster $1$ is ranked as the second least. 
\begin{table}[h]
\caption{Weight and rank of objectives for German credit dataset.}
\label{tab:german_objectives}
\begin{tabular}{lrr}
\toprule
\textbf{Objective} & \multicolumn{1}{l}{\textbf{Weight}} & \multicolumn{1}{l}{\textbf{Rank}} \\
\midrule
$\downarrow$VIF              & 0.20135 & 1 \\
$\downarrow$Subset size        & 0.17503 & 2 \\
$\uparrow$Equalised odds     & 0.15916 & 3 \\
$\uparrow$F1 Score           & 0.15542 & 4 \\
$\uparrow$Balanced accuracy  & 0.15536 & 5 \\
$\uparrow$Statistical parity & 0.15368 & 6\\
\bottomrule
\end{tabular}
\end{table}

Next, we derive the weights and rank the objectives based on their values in the solutions as explained in Section~\ref{experiments}. As shown in Table~\ref{tab:german_objectives}, the VIF objective has the highest weight and ranks first, while statistical parity ranks least. The ranks reflect the variance in the values of the objectives for the solutions.

Then, With these weights, we rank the solutions using the TOPSIS method. The solution with one feature -\textit{beszeit (Employment Duration)}- ranks the highest. This feature represents how long the account holder has been in his/her employment status. Comparing this to the baseline for ensuring fairness by eliminating the sensitive feature, Table~\ref{tab:german_topsis_baseline} shows again that it is preferable to include the fairness as an objective for feature selection than to eliminate it from the analysis.

\begin{table}[ht]
\caption{Comparison of TOPSIS top solution to baseline for German credit dataset.}
\label{tab:german_topsis_baseline}
\begin{tabular}{lrr}
\toprule
\textbf{Objective} & \multicolumn{1}{l}{\textbf{TOPSIS - 1}} & \multicolumn{1}{l}{\textbf{Baseline}} \\
\midrule
$\downarrow$Subset size        & \textbf{1}              & 19                     \\
$\uparrow$Balanced accuracy  & 0.5            & \textbf{0.6436}                  \\
$\uparrow$F1 Score           & 0.8235         & \textbf{0.8265}                   \\
$\downarrow$VIF              & \textbf{0}              & 13.7696                  \\
$\uparrow$Statistical parity & \textbf{1 }             & 0.7751                  \\
$\uparrow$Equalised odds     & \textbf{0.8000}         & 0.4144                  \\
\bottomrule
\end{tabular}
\end{table}

On the individual feature level, according to the mean SHAP values for the features selected at least once, \textit{laufkont (Status)} feature is the most relevant. This feature represents the status (i.e., amount of money) of the account holder's checking account. By frequency, \textit{moral (Credit History)} and \textit{beszeit (Employment Duration)} were most selected; $10$ out of $19$ times. 

\subsection{Sensitivity}
When using methods like TOPSIS, which rely on weights for solution ranking, obviously the weights impact the ranking obtained. Hence, we analyse how the chosen weights for the objectives impact the resulting ranking in our use cases. In Particular, we consider three cases: equal weights, $r/\sigma$, and entropy  \cite{oluah2020selection}, and for each case, we rank the solutions of the datasets using TOPSIS.  

\begin{table}[h]
\caption{Impact of objective weights on ranking of solutions.}
\label{tab:weigh_change}
\resizebox{\columnwidth}{!}{
\begin{tabular}{l|lll|lll}
\toprule
     & \multicolumn{3}{c|}{\textbf{Diabetes}} & \multicolumn{3}{c}{\textbf{German credit}} \\
     \midrule
Rank & Equal   & $r/\sigma$   & Entropy  & Equal     & $r/\sigma$    & Entropy    \\
\midrule
1    & soln4   & soln4  & soln9    & soln3     & soln3    & soln3      \\
2    & soln8   & soln3   & soln12   & soln2     & soln2    & soln2      \\
3    & soln3   & soln8   & soln5   & soln7     & soln7    & soln1      \\
4    & soln9   & soln9   & soln14   & soln6     & soln1    & soln7      \\
5    & soln10  & soln5   & soln23    & soln9     & soln9    & soln4     \\
\bottomrule
\end{tabular}}
\end{table}

Table~\ref{tab:weigh_change} shows the top 5 solutions for each dataset according to the various objective weights used. For the \textit{German credit} dataset, the top solutions chosen for all weights are similar and the highest ranked solution remains the same in all cases. However, for the \textit{Diabetes} dataset, changing the weights has a stronger impact on the ranking of solutions. The highest ranked solution in the three cases is not the same as well as other solutions that made it to the top $5$; especially for entropy weights. In general, the decision maker could consider different objective weights before choosing the most appropriate.

\section{Related Work} \label{related}
Many-objective optimisation are problems that require the simultaneous optimisation of four or more objectives. The goal is to find a diverse set of solutions that lie close to the true Pareto font \cite{chand2015evolutionary}. On the other hand, FS aims to find a subset of features from a dataset that are most relevant to the task at hand; a key part of this is defining relevance, which cannot be fully described by a single objective \cite{hicks2022evaluation}. Depending on the task, it requires considering up to four or more objectives referred to as Many-Objective Feature selection (MOFS). 

Using MOFS, Zhang et al. optimise five objectives for network anomaly detection in \cite{zhang2020many}, Pal et al. optimise five objectives for motor imagery actions prediction in \cite{pal2016many}, Dong et al. optimise seven objectives for multi-label classification in \cite{dong2020many}, and Li et al. optimise four objectives for large scale feature selection in in \cite{li2020dividing}. This shows the broad application of MOFS, however, how to choose a final solution from a large set non-dominated ones has not be studied in general.  

The mechanisms used to present and choose a final solution in previous work include: box plots  \cite{zhang2020many}, bar charts \cite{pal2016many}, scatter plots \cite{dong2020many}, parallel coordinate plots \cite{li2020dividing}, and tables  \cite{zhang2020many, pal2016many, dong2020many}. Additionally, measures such as the hypervolume indicator \cite{pal2016many} and inverse generation distance -which is the distance from the ideal solution vector- \cite{zhang2020many} are also used. However, these approaches ignore one or more viewpoints, especially the feature viewpoint which is unique to MOFS, since FS requires a consideration of not just the solutions (which depend on the objectives) but also the features included in those solutions.

In many-objective optimisation, providing decision support for the decision maker remains a challenge because the commonly used methods for presenting the solutions become inadequate with the growing number of dimensions needed. To remedy this, the dimensions can be reduced or multiple presentations used for completeness \cite{korhonen2008visualization}. However, the former comes with loss of information and the later requires clear guidelines.

It is therefore imperative to design a methodology that facilitates the interpretability of MOFS results holistically, taking into account all three viewpoints that make up the process- objectives, solutions, and features.

\section{Conclusion} \label{conclusion}
MOFS produces a potentially large set of solutions with completely different values for the objectives being considered. Thus, we have proposed a methodology to systematically analyse the results obtained by MOFS and presented its application to two concrete datasets that showcase its soundness and usefulness. The results of our experiments framed under the studies performed in the project DEDS (MSCA-ITN GA No 955895) 
allow to conclude that (1) FS plays a relevant role in building fair machine learning models as well as (2) how the relevance of an objective is problem dependent.

In the future, this work can be extended by empirically assessing the impact of various methods of generating weights for the objectives based on dataset characteristics and solutions. Furthermore, the methodology can be incorporated into existing or new FS packages to ensure accessibility. Finally, we will conduct a usability study to assess how well the methodology is interpreted by users as compared to current practices.
\section{Acknowledgements} \label{ack}
The project leading to this publication has received funding from the European Commission under the European Union’s Horizon 2020 research and innovation programme (grant agreement No. 955895). Moreover, A. Abelló and B. Bilalli are funded by the Spanish Ministerio de Ciencia e Innovación under project  PID2020-117191RB-I00, funding scheme AEI/10.13039/501100011033.

\bibliographystyle{ACM-Reference-Format}
\bibliography{literature}

\end{document}